\documentclass{article}
\usepackage{arxiv}
\usepackage[utf8]{inputenc} 
\usepackage[T1]{fontenc}    
\usepackage{hyperref}       
\usepackage{url}            
\usepackage{booktabs}       
\usepackage{amsfonts}       
\usepackage{nicefrac}       
\usepackage{microtype}      
\usepackage{lipsum}
\usepackage{graphicx}
\usepackage{amsmath}
\usepackage{algorithm}
\usepackage{algorithmic}
\usepackage{multirow}

\title{Discriminative Dimension Reduction\\based on Mutual Information}

\author{
 Orod Razeghi\\
  School of Computer Science\\
  University of Nottingham\\
  Nottingham, United Kingdom NG7 2RD\\
 \And
 Guoping Qiu\\
  School of Computer Science\\
  University of Nottingham\\
  Nottingham, United Kingdom NG7 2RD\\
}

\begin{document}
\maketitle

\begin{abstract}
The "curse of dimensionality" is a well-known problem in pattern recognition. A widely used approach to tackling the problem is a group of subspace methods, where the original features are projected onto a new space. The lower dimensional subspace is then used to approximate the original features for classification. However, most subspace methods were not originally developed for classification. We believe that direct adoption of these subspace methods for pattern classification should not be considered best practice. In this paper, we present a new information theory based algorithm for selecting subspaces, which can always result in superior performance over conventional methods. This paper makes the following main contributions: i) it improves a common practice widely used by practitioners in the field of pattern recognition, ii) it develops an information theory based technique for systematically selecting the subspaces that are discriminative and therefore are suitable for pattern recognition/classification purposes, iii) it presents extensive experimental results on a variety of computer vision and pattern recognition tasks to illustrate that the subspaces selected based on maximum mutual information criterion will always enhance performance regardless of the classification techniques used.
\end{abstract}

\keywords{Dimension Reduction \and Feature Selection \and Mutual Information \and Subspace Methods \and Classification}

\section{Introduction}

In many machine vision and pattern recognition applications, the infamous "curse of dimensionality" is a well-known problem. A widely used approach to alleviate this complication is subspace methods, where the original data is projected onto a new space in which lower dimensional feature vectors are used to approximate the original vectors. Amongst conventional subspace methods are: Principal Component Analysis (PCA) \cite{Jolliffe2002}, Linear Discriminant Analysis (LDA) \cite{FISHER1936}, various frequency analysis based transforms such as Fourier Transform (FT) \cite{Titchmarsh1937}, and its derivative Discrete Cosine Transform (DCT) \cite{Ahmed1974}, short-time Fourier Transform, Wavelet Transform (WT) \cite{Chui1992}, and other variations of frequency analysis method such as Hadamard Transform (HT) \cite{Beauchamp1984}. Random Projection (RP) \cite{Bingham2001}, where the original data is projected onto a lower dimensional random directions, has also been used for dimensionality reduction.

Nevertheless, it is reasonable to claim that all these subspace methods were not originally developed specifically for pattern recognition or object classification applications. PCA is for identifying subspaces, in which the input data has the largest variance, such that the inverse transform from a lower dimensional subspace recovers the original data with minimum loss of energy. FT, DCT, WT, and HT are all for retaining the lower frequency components of the original data. Surprisingly, pattern recognition literature conventionally adopts these methods, as they were originally developed for dimensionality reduction, without questioning if they also make theoretical and practical sense when applied to pattern recognition.

We believe an ideal representation is in a space where classes of data are well separable. As we will demonstrate later, directly applying these methods in pattern recognition or classification is not always the best practice, and a new information theory based method for selecting the subspaces can enhance the performance of a learning system substantially. This papers makes the following main contributions:

\begin{enumerate}
	\item It enhances a common practice widely used by practitioners in the field of pattern recognition. To the best of our knowledge, this is the first paper which highlights the interesting fact that in implementation of dimensionality reduction subspace methods, such as \cite{Fidler2006,Zhao2000,Yang2000,Haykin1998,Belhumeur1996} for pattern recognition or classification applications, practitioners should not directly adopt the conventional methods but instead explicitly opt for a discriminative subspace from the transforms.
	\item It develops an information theory based technique for systematically selecting subspaces that are discriminative and therefore are suitable for pattern recognition or classification purposes.
	\item It presents extensive experimental results on a variety of computer vision and pattern recognition tasks to illustrate that the subspaces selected based on maximum mutual information criterion will always improve performance regardless of the classification techniques in use.
\end{enumerate}


In the rest of this paper, our setup is the regular multiclass setting, where we have a labelled dataset $\{(x_i,y_i) \in X \times Y\}$ sampled $iid$ from a distribution $D$ on $\mathbb{R}^d \times [l]$. We therefore need a classifier $f : \mathbb{R}^d \rightarrow [l]$ with low generalisation error $\mathbb{P}_D(f(x) \neq y)$. To keep focus on the effectiveness of our subspace selection method, we restrict ourselves to three classifiers: Random Forest, Support Vector Machine, and Naive Bayes in the evaluation section of this paper. We believe our discriminative dimensionality reduction method can also improve the performance of other classifiers given any particular settings.

\section{Information Theory Background}

Our proposed method is close in nature to existing methods that work by finding suitable subspaces constructed from data. These methods generally find directions $v$ that maximise a signal to noise ratio:

\begin{equation}
	R(v) = \frac{v^TSv}{v^TNv}
\end{equation}

\noindent where matrices $S$ and $N$ are selected such that quadratic forms $v^TSv$ and $v^TNv$ represent signal and noise respectively along direction $v$. This defined ratio allows us to categorise similar methods in those that can produce many directions, and those that can generate discriminative directions. One of the most straightforward statistics involving both features and labels to extract directions is the matrix $\mathbb{E}[xy^T]$. This is the collection of class-conditional mean feature vectors in a multiclass classification setting. However, it is relatively safe to expect that such simple first moment statistics fails to contain all the information available in the data distribution. Alternatively, a collection of the conditional second moment matrices $C_i = \mathbb{E}[xx^T|y=i]$ can be used to extract features. Nevertheless, there is no reason to expect that these extracted directions are specific to class $i$. The directions may be very similar for all classes, and hence not very discriminative. A simple solution to this problem is to work with ratio of expected projection magnitudes conditional on different class labels. This necessitates to address which class pairs to extract directions. When the number of class labels is modest, it is possible to consider all ordered pairs of classes but unfortunately this is not the case in many applications.

We believe that it may be advantageous to explore higher than second order statistical information to derive a discriminative subspace, which not only enables low dimensional representation of inputs but also allows input projections to be well-separated. For instance, kernel based subspace methods \cite{Zhang2005,Schoelkopf1998,Kim2003} exploit higher order statistics to derive the subspace. Information theory \cite{Cover2012} can also be used to benefit from higher order statistics. Mutual information measures general statistical dependence between variables rather than their linear correlations. It is also invariant to monotonic transformations performed on the variables. These illustrate a number of advantages that information theoretic approaches may have over similar methods for deriving discriminative subspaces. For instance, in \cite{Vidal-Naquet2003,Qiu2008} mutual information is employed in deriving supervised but part-based representations for object classification.

Our proposed methodology based on mutual information exploits Fano's inequality \cite{Fano1961} in a similar manner as authors in \cite{Fisher1998,Butz2002}. 
We intend to test our introduced method on both data-independent, and data-dependent dimensionality reduction techniques and to the best of our knowledge our specific method of dimensionality reduction in this paper is novel. We prove that our solution empirically works well, as we illustrate in the evaluation section of this paper.

\section{Mutual Information Subspace Implementation}

Inspired by the aforementioned advantages of mutual information over alternative solutions, we introduce the implementation of a method based on information theory. This technique exploits mutual information to guard against selecting non-discriminative directions, while allowing the extraction of a diverse range of transformation vectors. Formally, we let $X$ and $Y$ be discrete random variables with sets of possible outcomes. We then define the mutual information between $X$ and $Y$ as:

\begin{equation} \label{eq:mi}
	I(X;Y) = \sum_{y \in Y} \sum_{x \in X} p(x,y) \log \left(\frac{p(x,y)}{p(x)p(y)}\right)
\end{equation}

\noindent where $p(x,y)$ is the joint probability distribution function of $X$ and $Y$, and $p(x)$ and $p(y)$ are the marginal probability distribution functions of $X$ and $Y$ respectively.

Mutual information measures the amount of information $x$ conveys about $y$. In the context of object classification, Fano's inequality \cite{Fano1961} provides us with a lower bound for the probability of error, an upper bound for the probability of correct classification. Formally, the probability of misclassification error $P_e = P(y \neq y')$ has the following bound:

\begin{equation} \label{eq:fano}
	P(y \neq y') >= \frac{ H(Y) - I(X;Y) - 1 }{ log(C) }
\end{equation}

\noindent where $H(Y)$ is the entropy of $Y$, $X$ is the ensemble of random variable $x$, and $C$ is the number of object classes.

The formula (\ref{eq:fano}) quantifies at best how well we can classify objects using features $x$. However, an upper bound of the probability of misclassification error cannot be expressed in terms of Shannon's entropy. The best one can do is to minimise the lower bound to ensure an appropriately designed classification algorithm performs well. Since both $C$ and $H(Y)$ are constants, we can maximise the mutual information $I(X;Y)$ to minimise the lower bound of the probability of misclassification error. At this point, the task develops into finding the transform function that minimises this lower bound. We therefore implement the preceding transform function by finding a low-dimensional representation $f$ of the original set of $N$ d-dimensional observations $X_{d \times N}$. This is achieved by projecting the original input data onto a k-dimensional $(k<<d)$ subspace using a $k \times d$ matrix $G$:

\begin{equation}
	f_{k \times N} = G_{k \times d}X_{d \times N}
\end{equation}

In this paper, we are motivated by mutual information and an information theoretic criterion to select the projection matrix $G^*$:

\begin{equation} \label{eq:maxim}
	G^* = \arg\max_{\forall G} I(GX;Y)
\end{equation}

\noindent where $Y$ is the identity variable of input variable $X$, $I(X;Y)$ is the mutual information between $X$ and $Y$.

The mutual information $ I(GX;Y)$ is calculated by estimating the probability density from a finite number of samples. Let us assume that we have a $N$ number of examples in the training set. The probability densities $p(x)$, $p(y)$, and $p(x,y)$ can be approximated by histograms. 
The difference between the true value $\overline{I}$ and the estimation $I$ of the mutual information can be estimated by adapting the analysis of \cite{Li1990}, as the following:

\begin{equation}
	\Delta I \equiv I - \overline{I} \approx \frac{1}{2N} \left( \sum_{x,y} \frac{(\delta n_{xy})^2}{n_{xy}} - \sum_x \frac{(\delta n_x)^2}{n_x} - \sum_y \frac{(\delta n_y)^2}{n_y} \right)
\end{equation}

\noindent where the sums are over the discretised intervals and $\delta n$ are the fluctuations of the countings with respect to the mean values $(\delta n = n - \overline{n})$. The approximation is valid up to the second order of the relative fluctuations, and if the ratios do not change significantly with $x$ and $y$.

Different subspace methods differ in their way of computing and selecting base vectors of the projection matrix $G$. We need to clarify that our criterion for selection of base vectors in the projection matrix $G$ differs intrinsically from their conventional counterparts. Specifically, we want to employ the maximum mutual information criterion (\ref{eq:maxim}) to select the appropriate $k$ base vectors.

To find the first base vector of $G$, we select one computed vector from a subspace method at a time, and project all other computed vectors from the training set onto that selected vector. The projections are a set of scalar numbers, which can be discretised. The samples' identities can be used to estimate the joint probability. The joint probability can be deployed to estimate the mutual information between the projections and the class distribution, as discussed previously. The vector with projection outputs that maximises the mutual information is selected as the first transform base of matrix $G$. This base is subsequently removed from the vectors' set.

The process continues until all required $k$ bases are found. If we have a large enough pool of samples, it is reasonable to assume that most informative representative bases will be selected. The representational quality and discriminative power of $f$ is dependent on the computed base vectors of matrix $G$. In the rest of this paper to clarify the practicality of this approach, we exemplify the computation and selection procedures of a data-independent, data-dependent, and the random projection methods of dimensionality reduction using mutual information criterion. The pseudo code in \ref{alg} summarises our process described in this section thus far.

\begin{algorithm} [htbp]
	\caption{Mutual Information Subspace Algorithm}
	\label{alg}
	\begin{algorithmic}
		\REQUIRE Observations: $X_{d \times N}$, Labels: $Y$, Number of base vectors: $k$
		\STATE Step 0: \parbox[t]{\dimexpr\linewidth-\algorithmicindent}{
			Compute subspace transformation matrix $G$\\
			e.g. eigenvectors of covariance matrix form $G_{d \times d}$ in PCA}
		\STATE Step 1: \parbox[t]{\dimexpr\linewidth-\algorithmicindent}{
			Compute projections of samples $X$ onto base vectors of $G$\\
			i.e. form projection matrix $Z_{d \times N}=GX$}
		\STATE Step 2: \parbox[t]{\dimexpr\linewidth-\algorithmicindent}{
			Compute mutual information for every base vector of $Z$ as in eq.\ref{eq:mi}\\
			i.e. calculate $I(Z_{i\times1};Y), \forall i\in d$}
		\STATE Step 3: \parbox[t]{\dimexpr\linewidth-\algorithmicindent}{
			Sort all base vectors based on their calculated $I(Z_{i\times1};Y)$\\
			i.e. construct matrix $G^* = \emptyset$
			\WHILE {there exist unsorted base vectors $v$}
				\STATE $v = \arg\max\limits_{\forall i\in d} I(Z_{i\times1};Y)$
				\STATE $G^* = G^* \cup \{v\}, Z = Z - \{v\}$
			\ENDWHILE}
		\RETURN First $k$ rows of $G^*$
	\end{algorithmic}
\end{algorithm}

\subsection{Examples of Common Subspace Methods}

\indent\textbf{Data Independent Transform - DCT:} The projection matrix $G$ in the Discrete Cosine Transform (DCT) method of dimensionality reduction is the transform coefficients. Conventionally, reduction is achieved in the inverse transform by discarding the transform coefficients corresponding to the highest frequencies. In contrast to the convention, we propose to use the mutual information criterion (\ref{eq:maxim}) to select the $k$ transform coefficients used in the projection matrix, and not simply the coefficients corresponding to the lowest frequencies.

\textbf{Data Dependent Transform - LDA:} LDA computes an optimal projection by minimising the within-class distance and maximising the between-class distance simultaneously, thus achieving maximum class discrimination. The optimal transformation in LDA can be readily computed by applying an eigendecomposition on the so-called scatter matrices. More specifically, eigenvectors corresponding to the $k-1$ largest eigenvalues form columns of $G$. Instead of relying on largest eigenvalues to form the projection matrix, our proposed mutual information criterion selects the base vectors of $G$.

\textbf{Data Dependent Transform - PCA:} In Principal Component Analysis (PCA), eigenvalue decomposition of data covariance matrix is computed as $\mathbb{E}\{XX^T\} = E \Lambda E^T$, where the columns of matrix $E$ are the eigenvectors of data covariance matrix $\mathbb{E}\{XX^T\}$ and $\Lambda$ is a diagonal matrix containing the respective eigenvalues. The $k$ eigenvectors corresponding to the $k$ largest eigenvalues of the covariance matrix form the projection matrix $G$. In contrast to this traditional approach of selecting the first $k$ vectors, we rely on the mutual information criterion (\ref{eq:maxim}) explained previously to select the appropriate bases of matrix $G$.

\textbf{Random Projection - RP:} A simple probability distribution can form the base vectors of the projection matrix $G$ in the Random Projection method of dimensionality reduction:

\begin{equation}
	g_{ij} = 
	\begin{cases}
		+1 & \text{with probability 1/3}\\
		 0 & \text{with probability 1/3}\\
		-1 & \text{with probability 1/3}
	\end{cases}
\end{equation}

Conventionally, the first $k$ computed vectors from this distribution constitute the projection matrix $G$. However as before, we propose to use the mutual information criterion (\ref{eq:maxim}) to select the required $k$ bases instead.

\subsection{Useful Properties}

The feature descriptors resulting from maximising equation (\ref{eq:mi}) have a number of useful properties that we list below:

\textbf{Proposition} (Maximum Dependence) By maximising equations (\ref{eq:mi}), we ensure that two random variables $X$ and $Y$ are statistically as dependent as possible. This means that feature vectors most relevant to a certain class is always preferred.

\textit{Proof.} Mutual information $I(X;Y) = 0$ if and only if $X$ and $Y$ are independent random variables. In such case the joint probability between the two variables is $p(x,y) = p(x)p(y)$, and therefore:

\begin{equation}
	\log{\left(\frac{p(x,y)}{p(x)p(y)}\right)} = \log{1} = 0
\end{equation}

This criterion enables our algorithm to maximally exploit the data by selecting the most informative descriptors of a certain class.

\textbf{Proposition} (Nonlinear Separations) Mutual information ability to consider nonlinear relations between variables can be advantageous over linear methods of analysis like correlation.

\textit{Proof.} Mutual information is capable of measuring general dependence between two variables. Variables $x$ and $y$ are linearly independent if $\mathbb{E}(xy) = \mathbb{E}(x)\mathbb{E}(y)$, and generally independent if $p(x,y) = p(x)p(y)$. Hence, general independence implies linear independence, but not vice versa. This property enables this algorithm to be advantageous over linear methods of analysis.

%

\section{Experiments and Results}

We evaluate our discriminative subspace selection method on a number of benchmark datasets. However, we firstly assess simple synthetic data to graphically illustrate the efficacy of our proposed technique. 
All results presented in this section are based on a 5-time repeated random sub-sampling cross validation method. We fix three commonly used multiclass classifiers and compare their outcomes to solely examine the performance of our algorithm and discard other potentially influential factors. The selected classifiers are an ensemble of 200 bagged decision trees, an RBF kernel SVM, and a naive Bayes model.

It is imperative to note that the main focus here is not to exactly achieve the-state-of-the-art classification performance on all datasets through a vigilant engineering procedure, but to emphasise the usefulness of our mutual information technique given any method of subspace dimensionality reduction. Our algorithm outperforms the original subspace methods in all tests or at minimum produces comparable results. The state-of-the-art performance is easily attainable once our feature selection approach is combined with carefully crafted descriptors and fine-tuned hyperparameters of robust classifiers.

\subsection{Synthetic Data}

In this pilot experiment, we generate a 100-by-2 matrix $R$ of random vectors chosen from a multivariate normal distribution with mean vector $\mu$, and a symmetric positive semi-definite covariance matrix $\Sigma$ to simplify the visualisation process. The synthetic data is an example of binary classification problem, where two sets of 50 instances belongs to 2 discriminable class labels. The data is randomly split in a 50:50 training and testing sets.

We project the 2-$d$ features in 1-$d$ subspace and use a naive Bayesian classifier to categorise the data in the projected 1-$d$ space. The purpose of this experiment is to verify the soundness of our method and to demonstrate the possible advantages of our technique over conventional subspace methods. We hence compare our algorithm with the well-known subspace method of Principal Component Analysis, and measure the mutual information between the projected 1-$d$ features and the data's class labels to examine the relation between mutual information and classification error.

\begin{figure} [htbp] \label{fig:toy}
	\centering
	\includegraphics[width=.75\linewidth]{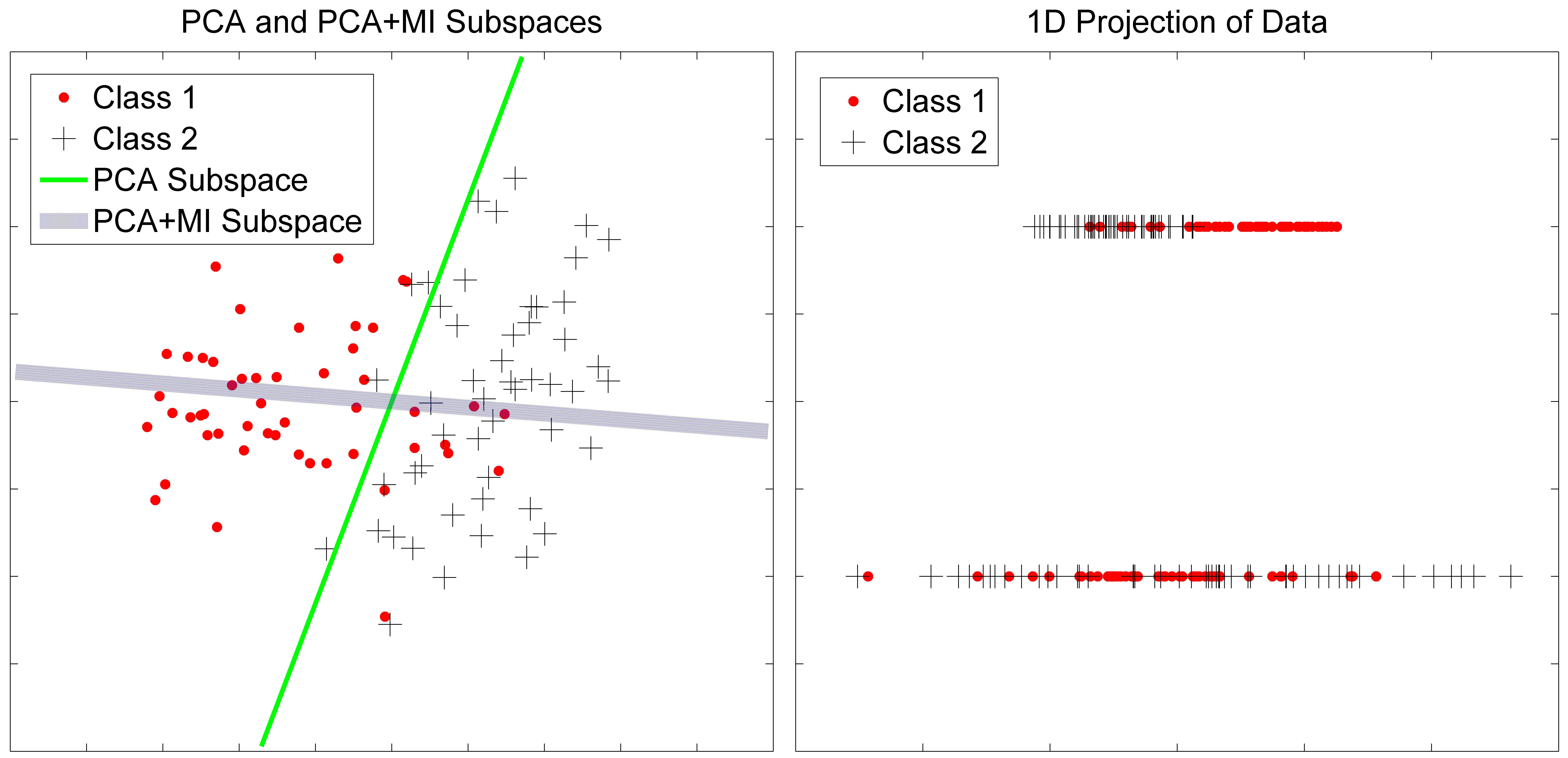}
	\caption{2D illustration of a two class dataset, where PCA and mutual information subspaces have been drawn. The projection on a 1D line clearly demonstrate the separability of data using our subspace selection method (right box, top line).}
\end{figure}

Figure \ref{fig:toy} depicts a situation, where projecting the data onto the 1-$d$ PCA base fails to discriminate between the two classes, whilst projecting the data onto our subspace alleviates the classification problem by making classes easier to separate. In this example, PCA finds the direction of maximal variance but fails to determine the most discriminative direction. From this simple experiment, we also conclude that the mutual information and classification error have a direct relation: the higher the mutual information contained in the subspace, the lower the classification errors are and vice versa.

Principal Component Analysis is a widely used linear transform for dimensionality reduction. It is an optimal reduction technique in the mean square error sense. The eigen subspace captures the directions of maximal variance in data. Nevertheless, as we just illustrated in Figure \ref{fig:toy}, the maximal variance and discriminative directions are not guaranteed to coincide. Therefore, PCA subspace is not always appropriate for representing the data in a classification settings. We believe that our algorithm captures higher order, more general statistical information, and therefore is a more suitable candidate than the alternative solutions.

\subsection{DERM2309 Skin Condition Dataset}

This dataset \cite{Razeghi2014} contains images of skin conditions from 44 different diseases. There are 880 training and 1429 testing images, totalling 2309 images in the dataset. In the original release of this dataset, there are 20 training images per class, and the rest are used for testing. The sheer difficulty of this dataset, in addition to our adherence to its original split of training and testing sets lead to the observed low average classification accuracies in our experiments. 
A full list of class labels in this dataset is available from \cite{Razeghi2014} but to mention a few: skin lesion images range from different types of Eczema to various cancerous conditions, such as Superficial Spreading Melanoma. 
The dataset also contains 37 possible questions that summarise the patient's skin lesion characteristics. 
Table \ref{tab:skinMI} represents the mean classification accuracies based on different fractions of data's original dimension. The mean accuracy of the same classifiers using no dimensionality reduction technique on this dataset saturates at 20.22\%. These results are based on PHOW-HSV, and textual features enclosed in the public release of this dataset.

\begin{table} [htbp]
\centering
\caption{Mean Accuracies in Percentage on DERM2309 Dataset} \label{tab:skinMI}
\footnotesize{
	\begin{tabular}{@{}l@{ } | l@{ }l@{ }l@{ }l@{ } | l@{ }l@{ }l@{ }l@{ } | l@{ }l@{ }l@{ }l@{}}
		\hline
		Classifier & \multicolumn{4}{c}{Random Forest} & \multicolumn{4}{c}{SVM}         & \multicolumn{4}{c}{Naive Bayes}\\
		\hline
		Dimension  & 10\%  & 30\%  & 50\%  & 70\%      & 10\%  & 30\%  & 50\%  & 70\%    & 10\%  & 30\%  & 50\%  & 70\%\\
		\hline\hline
		DCT        & 16.52 & 16.03 & 16.79 & 15.89     & 14.75 & 14.75 & 15.75 & 15.75   & 12.27 & 12.97 & 11.99 & 11.64\\
		DCT+MI     & 16.66 & 17.49 & 16.79 & 17.91     & 16.75 & 16.75 & 17.75 & 17.75   & 14.38 & 14.45 & 13.38 & 13.45\\\hline
		LDA        & 12.06 & 15.12 & 20.90 & 25.40     & 14.84 & 15.32 & 19.26 & 20.23   & 10.96 & 12.67 & 16.74 & 20.41\\
		LDA+MI     & 15.21 & 19.74 & 23.97 & 26.87     & 17.43 & 19.01 & 19.75 & 20.23   & 14.25 & 17.67 & 19.32 & 20.41\\\hline
		PCA        & 13.37 & 20.64 & 18.54 & 16.72     & 17.06 & 19.65 & 18.25 & 17.27   & 12.32 & 15.52 & 16.98 & 17.00\\
		PCA+MI     & 15.26 & 21.20 & 19.17 & 18.61     & 19.30 & 19.86 & 18.25 & 17.34   & 13.04 & 17.01 & 17.84 & 17.07\\\hline
		RND        & 11.24 & 12.08 & 12.43 & 13.06     & 16.17 & 15.75 & 15.75 & 15.75   & 10.22 & 12.25 & 14.83 & 15.81\\
		RND+MI     & 14.46 & 14.60 & 14.46 & 15.44     & 16.17 & 15.82 & 15.75 & 15.75   & 13.43 & 14.76 & 15.81 & 15.95\\
		\hline
	\end{tabular}
}
\end{table}

\subsection{Yale Face Recognition Dataset}

The Yale Face database \cite{Belhumeur1996} contains 165 grayscale images of 15 individuals in GIF format. There are 11 images per subject, one per different facial expression or configuration: center-light, w/glasses, happy, left-light, w/no glasses, normal, right-light, sad, sleepy, surprised, and wink. Table \ref{tab:yaleMI} illustrates the mean classification accuracies based on different fractions of data's original dimension. The mean accuracy of the same classifiers using no dimensionality reduction method on this dataset levels at 86.66\%. The only visual feature used is image intensity values ranging from 0 to 255.

\begin{table} [htbp]
\centering
\caption{Mean Accuracies in Percentage on Yale Face Dataset} \label{tab:yaleMI}
\footnotesize{
	\begin{tabular}{@{}l@{ } | l@{ }l@{ }l@{ }l@{ } | l@{ }l@{ }l@{ }l@{ } | l@{ }l@{ }l@{ }l@{}}
		\hline
		Classifier & \multicolumn{4}{c}{Random Forest} & \multicolumn{4}{c}{SVM}         & \multicolumn{4}{c}{Naive Bayes}\\
		\hline
		Dimension  & 10\%  & 30\%  & 50\%  & 70\%      & 10\%  & 30\%  & 50\%  & 70\%    & 10\%  & 30\%  & 50\%  & 70\%\\
		\hline\hline
		DCT        & 64.44 & 66.67 & 68.89 & 68.89     & 67.78 & 65.56 & 65.56 & 65.56   & 37.78 & 51.11 & 51.11 & 46.67\\
		DCT+MI     & 68.89 & 68.89 & 68.89 & 68.89     & 76.67 & 67.78 & 65.56 & 65.56   & 71.11 & 60.00 & 51.11 & 46.67\\\hline
		LDA        & 80.00 & 80.00 & 80.00 & 80.00     & 80.00 & 80.00 & 80.00 & 80.00   & 80.00 & 80.00 & 80.00 & 80.00\\
		LDA+MI     & 82.22 & 82.22 & 82.22 & 80.00     & 82.22 & 82.22 & 82.22 & 80.00   & 82.22 & 82.22 & 82.22 & 80.00\\\hline
		PCA        & 82.22 & 77.78 & 68.89 & 53.33     & 75.56 & 73.33 & 71.33 & 71.33   & 62.22 & 82.22 & 71.11 & 64.44\\
		PCA+MI     & 88.89 & 82.22 & 80.00 & 68.89     & 79.56 & 77.33 & 71.33 & 71.33   & 66.67 & 86.67 & 75.56 & 71.11\\\hline
		RND        & 75.56 & 80.00 & 75.56 & 75.56     & 71.11 & 74.44 & 70.00 & 70.00   & 71.11 & 73.33 & 73.33 & 73.33\\
		RND+MI     & 77.78 & 82.22 & 77.78 & 77.78     & 71.11 & 76.67 & 70.00 & 70.00   & 73.33 & 73.33 & 77.78 & 75.56\\
		\hline
	\end{tabular}
}
\end{table}

\subsection{Oxford Flower Recognition Dataset}

The Oxford flowers dataset \cite{Nilsback2008} contains 17 different types of flowers. Each class contains 80 samples, 40 for training, 20 for validation, and the rest for testing. Table \ref{tab:flowerMI} describes the mean classification accuracies based on different fractions of data's original dimension. The mean accuracy of the same classifiers using no dimensionality reduction technique on this dataset remains about 45.17\%. The visual features used are: HSV colour histograms, SIFT, and MR8 texture descriptors.

\begin{table} [htbp]
\centering
\caption{Mean Accuracies in Percentage on Oxford Flower Dataset} \label{tab:flowerMI}
\footnotesize{
	\begin{tabular}{@{}l@{ } | l@{ }l@{ }l@{ }l@{ } | l@{ }l@{ }l@{ }l@{ } | l@{ }l@{ }l@{ }l@{}}
		\hline
		Classifier & \multicolumn{4}{c}{Random Forest} & \multicolumn{4}{c}{SVM}         & \multicolumn{4}{c}{Naive Bayes}\\
		\hline
		Dimension  & 10\%  & 30\%  & 50\%  & 70\%      & 10\%  & 30\%  & 50\%  & 70\%    & 10\%  & 30\%  & 50\%  & 70\%\\
		\hline\hline
		DCT        & 41.76 & 43.53 & 42.65 & 42.94     & 41.76 & 39.41 & 39.71 & 39.41   & 31.18 & 29.71 & 29.12 & 28.24\\
		DCT+MI     & 45.00 & 43.53 & 42.94 & 46.18     & 41.76 & 39.41 & 39.71 & 39.41   & 33.12 & 34.29 & 35.76 & 36.06\\\hline
		LDA        & 37.94 & 45.88 & 47.65 & 49.71     & 40.59 & 47.94 & 48.53 & 50.00   & 31.76 & 38.24 & 38.82 & 40.00\\
		LDA+MI     & 39.12 & 46.47 & 48.82 & 50.00     & 40.59 & 47.94 & 50.00 & 50.00   & 31.76 & 38.24 & 40.00 & 40.00\\\hline
		PCA        & 44.71 & 40.59 & 35.29 & 32.94     & 46.76 & 46.76 & 45.88 & 45.88   & 47.65 & 42.35 & 36.18 & 39.71\\
		PCA+MI     & 47.94 & 42.06 & 41.18 & 35.00     & 47.06 & 48.76 & 47.88 & 47.88   & 48.24 & 43.53 & 36.18 & 39.71\\\hline
		RND        & 38.82 & 40.88 & 39.41 & 42.06     & 44.12 & 40.88 & 39.12 & 40.88   & 35.88 & 39.41 & 40.59 & 42.35\\
		RND+MI     & 41.18 & 43.53 & 47.65 & 42.94     & 44.41 & 40.88 & 39.71 & 41.18   & 36.76 & 40.00 & 41.47 & 42.65\\
		\hline
	\end{tabular}
}
\end{table}

\subsection{Pascal VOC2007 Challenge Dataset}

Pascal visual object classes of 2007 challenge dataset \cite{Everingham2007} has 20 distinguishable classes, as follows: person, bird, cat, cow, dog, horse, sheep, aeroplane, bicycle, boat, bus, car, motorbike, train, bottle, chair, dining table, potted plant, sofa, and tv/monitor. Train, validation, and test sets have 9963 images in total containing 24640 annotated objects. Table \ref{tab:pascalMI} displays the average classification accuracies based on different fractions of original data. The mean accuracy of the same classifiers using no dimensionality reduction method on this dataset saturates at 30.36\%. The results are based on 15 publicly released visual descriptors: Gist, DenseSift, DenseSiftV3H1, HarrisSift, HarrisSiftV3H1, DenseHue, DenseHueV3H1, HarrisHue, HarrisHueV3H1, Rgb, RgbV3H1, Lab, LabV3H1, Hsv, and HsvV3H1.

\begin{table} [htbp]
\centering
\caption{Mean Accuracies in Percentage on Pascal VOC2007 Dataset} \label{tab:pascalMI}
\footnotesize{
	\begin{tabular}{@{}l@{ } | l@{ }l@{ }l@{ }l@{ } | l@{ }l@{ }l@{ }l@{ } | l@{ }l@{ }l@{ }l@{}}
		\hline
		Classifier & \multicolumn{4}{c}{Random Forest} & \multicolumn{4}{c}{SVM}         & \multicolumn{4}{c}{Naive Bayes}\\
		\hline
		Dimension  & 10\%  & 30\%  & 50\%  & 70\%      & 10\%  & 30\%  & 50\%  & 70\%    & 10\%  & 30\%  & 50\%  & 70\%\\
		\hline\hline
		DCT        & 32.45 & 32.33 & 32.18 & 32.96     & 30.50 & 30.96 & 32.49 & 30.96   & 25.37 & 26.31 & 26.24 & 26.62\\
		DCT+MI     & 34.24 & 33.55 & 33.71 & 33.49     & 33.20 & 31.55 & 33.49 & 33.18   & 27.72 & 27.03 & 27.71 & 27.90\\\hline
		LDA        & 25.61 & 26.96 & 28.02 & 30.08     & 26.55 & 28.31 & 30.90 & 32.80   & 23.43 & 25.55 & 28.78 & 29.24\\
		LDA+MI     & 27.33 & 27.96 & 30.50 & 33.99     & 29.72 & 30.83 & 31.55 & 33.55   & 25.50 & 28.18 & 28.78 & 29.24\\\hline
		PCA        & 32.77 & 32.74 & 30.96 & 30.83     & 30.78 & 31.46 & 30.02 & 31.96   & 29.53 & 28.94 & 28.06 & 27.99\\
		PCA+MI     & 34.78 & 34.49 & 33.46 & 32.08     & 32.80 & 34.49 & 33.96 & 33.49   & 29.72 & 29.22 & 28.25 & 28.28\\\hline
		RP         & 21.42 & 21.92 & 28.61 & 22.02     & 20.77 & 22.33 & 28.18 & 21.24   & 20.94 & 21.12 & 21.12 & 20.84\\
		RP+MI      & 22.80 & 23.43 & 29.43 & 25.30     & 21.18 & 24.55 & 29.71 & 24.08   & 22.66 & 22.22 & 21.66 & 21.66\\
		\hline
	\end{tabular}
}
\end{table}

\subsection{MSRC 21-class Dataset}

MSRC 21-class is a well-known dataset \cite{Shotton2006} that contains 591 images. Each image has pixel-level ground-truth labels from 21 semantic classes. These 591 images are split into 276 for training, 59 for validation, and the remaining 256 images for testing purposes. Table \ref{tab:msrc21MI} represents the mean classification accuracies based on different fractions as before. The mean accuracy of the same classifiers using no dimensionality reduction technique on this dataset levels about 72.25\%. These results are based on Texton, colour histograms, and PHOG visual features.

\begin{table} [htbp]
\centering
\caption{Mean Accuracies in Percentage on MSRC 21-class Dataset} \label{tab:msrc21MI}
\footnotesize{
	\begin{tabular}{@{}l@{ } | l@{ }l@{ }l@{ }l@{ } | l@{ }l@{ }l@{ }l@{ } | l@{ }l@{ }l@{ }l@{}}
		\hline
		Classifier & \multicolumn{4}{c}{Random Forest} & \multicolumn{4}{c}{SVM}         & \multicolumn{4}{c}{Naive Bayes}\\
		\hline
		Dimension  & 10\%  & 30\%  & 50\%  & 70\%      & 10\%  & 30\%  & 50\%  & 70\%    & 10\%  & 30\%  & 50\%  & 70\%\\
		\hline\hline
		DCT        & 51.94 & 56.13 & 57.74 & 57.10     & 52.26 & 52.26 & 56.26 & 56.26   & 40.32 & 39.68 & 41.94 & 44.19\\
		DCT+MI     & 54.84 & 56.13 & 58.06 & 59.35     & 54.26 & 54.26 & 59.26 & 59.26   & 42.58 & 46.45 & 47.42 & 47.10\\\hline
		LDA        & 50.00 & 50.00 & 50.00 & 50.00     & 50.00 & 50.00 & 50.00 & 50.00   & 41.29 & 46.77 & 46.77 & 46.77\\
		LDA+MI     & 51.94 & 54.19 & 52.26 & 50.00     & 51.94 & 54.84 & 52.90 & 50.00   & 41.29 & 50.32 & 50.32 & 46.77\\\hline
		PCA        & 56.77 & 53.55 & 46.13 & 46.77     & 50.65 & 49.03 & 49.68 & 50.00   & 42.58 & 47.74 & 46.13 & 40.97\\
		PCA+MI     & 60.32 & 57.74 & 54.52 & 47.42     & 52.26 & 49.35 & 49.68 & 50.00   & 43.23 & 48.06 & 46.45 & 41.94\\\hline
		RND        & 50.00 & 50.97 & 52.90 & 54.19     & 52.26 & 52.26 & 52.26 & 52.26   & 37.10 & 42.58 & 42.90 & 43.55\\
		RND+MI     & 54.19 & 58.06 & 56.45 & 59.68     & 58.26 & 58.26 & 58.26 & 58.26   & 40.65 & 42.90 & 45.48 & 45.81\\
		\hline
	\end{tabular}
}
\end{table}

\subsection{UCI Machine Learning Repository Datasets}

We further present experiments on two UCI repository datasets: UCI-Sonar and MFeat sets \cite{Bache2013}. Table \ref{tab:sonarMI} describes the mean classification accuracies of UCI-Sonar dataset based on different fractions of original dimension. The mean accuracy of the same classifiers using no dimensionality reduction on this dataset levels at 77.27\%. There is only one feature used for the purpose of classification.

\begin{table} [htbp]
\centering
\caption{Mean Accuracies in Percentage on UCI-Sonar Dataset} \label{tab:sonarMI}
\footnotesize{
	\begin{tabular}{@{}l@{ } | l@{ }l@{ }l@{ }l@{ } | l@{ }l@{ }l@{ }l@{ } | l@{ }l@{ }l@{ }l@{}}
		\hline
		Classifier & \multicolumn{4}{c}{Random Forest} & \multicolumn{4}{c}{SVM}         & \multicolumn{4}{c}{Naive Bayes}\\
		\hline
		Dimension  & 10\%  & 30\%  & 50\%  & 70\%      & 10\%  & 30\%  & 50\%  & 70\%    & 10\%  & 30\%  & 50\%  & 70\%\\
		\hline\hline
		DCT        & 79.55 & 82.95 & 82.95 & 81.82     & 77.73 & 77.73 & 77.73 & 77.73   & 61.36 & 57.95 & 56.82 & 56.82\\
		DCT+MI     & 81.82 & 82.95 & 82.95 & 88.64     & 81.73 & 81.73 & 81.73 & 81.73   & 62.50 & 60.23 & 60.23 & 56.82\\\hline
		LDA        & 87.50 & 87.50 & 87.50 & 87.50     & 87.50 & 87.50 & 87.50 & 87.50   & 83.77 & 83.77 & 85.77 & 85.77\\
		LDA+MI     & 89.77 & 87.50 & 87.50 & 87.50     & 87.50 & 87.50 & 87.50 & 87.50   & 87.77 & 87.77 & 89.77 & 89.77\\\hline
		PCA        & 76.14 & 77.27 & 75.00 & 75.00     & 79.55 & 81.82 & 77.27 & 79.55   & 73.86 & 68.18 & 67.05 & 69.32\\
		PCA+MI     & 82.95 & 82.95 & 78.41 & 76.14     & 80.68 & 82.95 & 79.55 & 80.68   & 76.14 & 71.59 & 68.18 & 70.45\\\hline
		RND        & 54.55 & 72.73 & 73.86 & 77.27     & 47.73 & 67.73 & 73.86 & 73.86   & 54.55 & 53.41 & 59.09 & 60.23\\
		RND+MI     & 76.14 & 80.68 & 80.68 & 81.82     & 57.95 & 67.73 & 77.73 & 77.43   & 75.00 & 75.00 & 73.86 & 76.14\\
		\hline
	\end{tabular}
}
\end{table}

MFeat contains 2000 handwritten numerals from 0 to 9. Table \ref{tab:mfeatMI} also illustrates the mean classification accuracies of MFeat dataset based on different fractions of data's original dimension. The mean accuracy of the same classifiers using no dimensionality reduction method on this dataset stands at 95.75\%. These classification results are based on 6 features publicly available.

\begin{table} [htbp]
\centering
\caption{Mean Accuracies in Percentage on UCI MFeat Dataset} \label{tab:mfeatMI}
\footnotesize{
	\begin{tabular}{@{}l@{ } | l@{ }l@{ }l@{ }l@{ } | l@{ }l@{ }l@{ }l@{ } | l@{ }l@{ }l@{ }l@{}}
		\hline
		Classifier & \multicolumn{4}{c}{Random Forest} & \multicolumn{4}{c}{SVM}         & \multicolumn{4}{c}{Naive Bayes}\\
		\hline
		Dimension  & 10\%  & 30\%  & 50\%  & 70\%      & 10\%  & 30\%  & 50\%  & 70\%    & 10\%  & 30\%  & 50\%  & 70\%\\
		\hline\hline
		DCT        & 96.25 & 96.50 & 97.00 & 97.00     & 90.75 & 90.75 & 93.75 & 93.75   & 86.75 & 88.50 & 89.25 & 89.25\\
		DCT+MI     & 96.75 & 96.75 & 97.00 & 97.00     & 91.75 & 91.75 & 95.75 & 95.75   & 90.50 & 90.75 & 91.00 & 91.75\\\hline
		LDA        & 61.25 & 95.00 & 97.00 & 98.00     & 64.50 & 95.75 & 98.00 & 98.00   & 65.25 & 96.25 & 97.00 & 98.00\\
		LDA+MI     & 65.75 & 95.25 & 97.00 & 98.00     & 71.75 & 96.25 & 98.00 & 98.00   & 76.50 & 97.50 & 98.00 & 98.00\\\hline
		PCA        & 97.25 & 97.00 & 96.75 & 95.00     & 79.75 & 88.25 & 94.50 & 95.00   & 96.75 & 95.50 & 95.00 & 94.00\\
		PCA+MI     & 97.50 & 97.50 & 97.00 & 97.50     & 79.75 & 91.25 & 94.50 & 98.00   & 96.75 & 96.00 & 95.25 & 94.00\\\hline
		RND        & 94.00 & 96.00 & 97.00 & 96.50     & 52.75 & 80.75 & 90.75 & 90.75   & 93.75 & 95.00 & 95.50 & 95.00\\
		RND+MI     & 96.50 & 98.00 & 97.50 & 98.50     & 54.50 & 80.75 & 96.75 & 96.75   & 93.75 & 95.75 & 95.50 & 96.25\\
		\hline
	\end{tabular}
}
\end{table}

\subsection{Interpretation of Results}

We believe our proposed method can improve the performance of any subspace procedure. The computational expense of DCT is $O(dN\log_2(dN))$. LDA's $O(d2N)$ calculation is dominated by the computation of the within-class scatter and its inverse. PCA is estimated as $O(d^2N)+O(d^3)$, and Random Projection complexity is $O(dkN)$. We know that the computational overhead from our algorithm on these subspace methods is negligible. It is not feasible to interpret a certain message from the results of random projection technique on our datasets, apart from a solid improvement over the original method. However, a few interesting points can be observed from the results on the DCT, LDA, and PCA procedures:

\textbf{First Few Low Frequency Bases are not Necessarily Discriminative:} The results of our DCT experiments on the selected datasets illustrate the point that the first few low frequency bases are not necessarily preferred by our mutual information method. Interestingly, 30.66\% of the first 10 frequencies selected by our mutual information algorithm is different from the original DCT technique.

\textbf{First Few Principal Components are Discriminative:} Unlike DCT, the results from all datasets in our experiments highlight the fact that the first 10 principal components are also usually selected by our mutual information method. There is only a 3.33\% difference between the selection of our mutual information method, and the original PCA technique. The same conclusion cannot be drawn from the largest eigenvalues and their corresponding eigenvectors returned by the LDA approach. There appears to be a significant difference between typical LDA selections and our proposed mutual information method.

\textbf{Performance of Algorithms Exhibit Asymptotic Behaviour at 70\%:} As the number of base vectors in the projection matrices increases, i.e. the dimension of input data escalates, the difference in classification accuracies of our mutual information method and the original subspace techniques becomes expectedly negligible. This behaviour starts to surface for DCT, LDA and PCA procedures with approximately 70\% of original dimension. Table \ref{tab:diff} displays these asymptotic results by listing the number of different bases returned by the two methods in percentage.

\begin{table} [htbp]
\centering
\caption{Difference in Bases returned by MI and the Conventional Methods in Percentage} \label{tab:diff}
\scriptsize{
	\begin{tabular}{@{}l l | l l l l l l l | l@{}}
	    \hline
	                         & Dim. & DERM  & Face  & Flower& VOC07 & MSRC  & Sonar & MFeat & Average\\
	    \hline\hline
	    \multirow{4}{*}{DCT} & 10\% & 78.30 & 83.33 & 55.45 & 37.16 & 87.72 & 66.67 & 48.44 & 65.29\\
	  	                     & 30\% & 63.21 & 60.53 & 52.42 & 32.55 & 69.19 & 61.11 & 43.81 & 54.68\\
	                         & 50\% & 44.15 & 34.38 & 41.09 & 26.75 & 54.36 & 43.33 & 40.43 & 40.64\\
	                         & 70\% & 26.01 & 22.47 & 26.62 & 17.96 & 31.67 & 26.19 & 26.65 & 25.36\\\hline
	    \multirow{4}{*}{LDA} & 10\% & 20.00 & 14.29 & 06.39 & 13.16 & 10.00 & 06.00 & 100.0 & 24.26\\
	                         & 30\% & 38.46 & 14.29 & 05.09 & 11.75 & 33.33 & 12.40 & 33.33 & 21.23\\
	                         & 50\% & 09.09 & 14.29 & 12.50 & 12.34 & 28.57 & 12.40 & 02.33 & 13.07\\
	                         & 70\% & 03.09 & 00.00 & 03.00 & 10.02 & 00.00 & 04.01 & 00.00 & 02.87\\\hline
	    \multirow{4}{*}{PCA} & 10\% & 12.83 & 11.96 & 12.75 & 12.97 & 16.14 & 20.00 & 04.50 & 13.02\\
	                         & 30\% & 11.89 & 08.56 & 11.52 & 13.02 & 11.45 & 17.65 & 08.19 & 11.75\\
	                         & 50\% & 06.57 & 09.33 & 10.73 & 10.34 & 13.52 & 03.45 & 04.85 & 08.39\\
	                         & 70\% & 05.54 & 10.33 & 10.39 & 04.71 & 09.77 & 04.88 & 08.52 & 07.73\\
	    \hline
	\end{tabular}
}
\end{table}

\section{Conclusion}

In this paper, we illustrated a novel method based on mutual information for discriminative subspace selection. We demonstrated empirical efficacy via multiple experiments on different datasets. Due to this empirical, computational, and statistical properties, we believe our proposed model has the potential capacity to be employed in a wide range of computer vision and pattern recognition problems.

\bibliographystyle{unsrt}
\bibliography{manuscript}  

\begin{thebibliography}{10}

\bibitem{Jolliffe2002}
Ian Jolliffe.
\newblock {\em Principal Component Analysis}.
\newblock Springer Series in Statistics. Springer, 2002.

\bibitem{FISHER1936}
Ronald~Aylmer Fisher.
\newblock The use of multiple measurements in taxonomic problems.
\newblock {\em Annals of Eugenics}, 7(2):179--188, 1936.

\bibitem{Titchmarsh1937}
Edward~Charles Titchmarsh.
\newblock {\em Introduction to the Theory of Fourier Integrals}.
\newblock Clarendon Press, 1937.

\bibitem{Ahmed1974}
N.~Ahmed, T.~Natarajan, and K.R. Rao.
\newblock Discrete cosine transform.
\newblock {\em IEEE Transactions on Computers}, C-23(1):90--93, January 1974.

\bibitem{Chui1992}
Charles~K. Chui.
\newblock {\em An Introduction to Wavelets}.
\newblock Wavelet analysis and its applications. Academic Press, 1992.

\bibitem{Beauchamp1984}
K.G. Beauchamp.
\newblock {\em Applications of Walsh and related functions, with an
  introduction to sequency theory}.
\newblock Microelectronics and signal processing. Academic Press, 1984.

\bibitem{Bingham2001}
Ella Bingham and Heikki Mannila.
\newblock Random projection in dimensionality reduction: applications to image
  and text data.
\newblock In {\em Knowledge Discovery and Data Mining}, pages 245--250, 2001.

\bibitem{Fidler2006}
S.~Fidler, D.~Skocaj, and A.~Leonardis.
\newblock Combining reconstructive and discriminative subspace methods for
  robust classification and regression by subsampling.
\newblock {\em IEEE Transactions on Pattern Analysis and Machine Intelligence
  (PAMI)}, 28(3):337--350, March 2006.

\bibitem{Zhao2000}
Wenyi Zhao.
\newblock Discriminant component analysis for face recognition.
\newblock In {\em 15th International Conference on Pattern Recognition},
  volume~2, pages 818--821, 2000.

\bibitem{Yang2000}
Ming-Hsuan Yang, N.~Abuja, and D.~Kriegman.
\newblock Face detection using mixtures of linear subspaces.
\newblock In {\em Fourth IEEE International Conference on Automatic Face and
  Gesture Recognition}, pages 70--76, 2000.

\bibitem{Haykin1998}
Simon Haykin.
\newblock {\em Neural Networks: A Comprehensive Foundation}.
\newblock Prentice Hall PTR, Upper Saddle River, NJ, USA, 2nd edition, 1998.

\bibitem{Belhumeur1996}
Peter~N. Belhumeur, João~P. Hespanha, Jo~ao~P.~Hespanha, and David~J.
  Kriegman.
\newblock Eigenfaces vs. fisherfaces: Recognition using class specific linear
  projection.
\newblock {\em IEEE Transactions on Pattern Analysis and Machine Intelligence
  (PAMI)}, 19:711--720, 1996.

\bibitem{Zhang2005}
Peng Zhang, Jing Peng, and C.~Domeniconi.
\newblock Kernel pooled local subspaces for classification.
\newblock {\em IEEE Transactions on Systems, Man, and Cybernetics, Part B:
  Cybernetics}, 35(3):489--502, June 2005.

\bibitem{Schoelkopf1998}
Bernhard Schölkopf, Alexander Smola, Er~Smola, and Klaus-Robert Müller.
\newblock Nonlinear component analysis as a kernel eigenvalue problem.
\newblock {\em Neural Computation}, 10:1299--1319, 1998.

\bibitem{Kim2003}
Sang-Woon Kim and B.John Oommen.
\newblock On using prototype reduction schemes and classifier fusion strategies
  to optimize kernel-based nonlinear subspace methods.
\newblock In Tamas (Tom)~Domonkos Gedeon and LanceChunChe Fung, editors, {\em
  AI 2003: Advances in Artificial Intelligence}, volume 2903 of {\em Lecture
  Notes in Computer Science}, pages 783--795. Springer Berlin Heidelberg, 2003.

\bibitem{Cover2012}
Thomas~M Cover and Joy~A Thomas.
\newblock {\em Elements of information theory}.
\newblock John Wiley \& Sons, 2012.

\bibitem{Vidal-Naquet2003}
M.~Vidal-Naquet and S.~Ullman.
\newblock Object recognition with informative features and linear
  classification.
\newblock In {\em Ninth IEEE International Conference on Computer Vision
  (ICCV)}, volume~1, pages 281--288, October 2003.

\bibitem{Qiu2008}
Guoping Qiu and Jianzhong Fang.
\newblock Classification in an informative sample subspace.
\newblock {\em Pattern Recognition}, 41(3):949--960, 2008.
\newblock Part Special issue: Feature Generation and Machine Learning for
  Robust Multimodal Biometrics.

\bibitem{Fano1961}
Robert~M Fano.
\newblock {\em Transmission of Information: A Statistical Theory of
  Communications}.
\newblock MIT Press Classics. M.I.T. Press, 1961.

\bibitem{Fisher1998}
J.W. Fisher and J.C. Principe.
\newblock A methodology for information theoretic feature extraction.
\newblock In {\em IEEE International Joint Conference on Neural Networks, IEEE
  World Congress on Computational Intelligence}, volume~3, pages 1712--1716,
  May 1998.

\bibitem{Butz2002}
Torsten Butz and Jean-Philippe Thiran.
\newblock Multi-modal signal processing: An information theoretical framework.
\newblock Technical report, Signal Processing Institute (ITS), Swiss Federal
  Institute of Technology (EPFL), 2002.

\bibitem{Li1990}
Wentian Li.
\newblock Mutual information functions versus correlation functions.
\newblock {\em Journal of Statistical Physics}, 60:823--837, 1990.

\bibitem{Razeghi2014}
Orod Razeghi and Guoping Qiu.
\newblock 2309 skin conditions and crowd-sourced high-level knowledge dataset
  for building a computer aided diagnosis system.
\newblock ISBI, 2014.

\bibitem{Nilsback2008}
M-E. Nilsback and A.~Zisserman.
\newblock Automated flower classification over a large number of classes.
\newblock In {\em The Indian Conference on Computer Vision, Graphics and Image
  Processing}, December 2008.

\bibitem{Everingham2007}
M.~Everingham, L.~Van~Gool, C.~K.~I. Williams, J.~Winn, and A.~Zisserman.
\newblock The pascal visual object classes challenge 2007 results (voc2007).
\newblock
  http://www.pascal-network.org/challenges/VOC/voc2007/workshop/index.html,
  2007.

\bibitem{Shotton2006}
J~Shotton, J~Winn, C~Rother, and Criminisi.
\newblock Msrc 21 class database.
\newblock http://jamie.shotton.org/work/data.html, 2006.

\bibitem{Bache2013}
K.~Bache and M.~Lichman.
\newblock {UCI} machine learning repository, 2013.

\end{thebibliography}

\end{document}